\documentclass[sigconf,natbib=true,authorversion=true]{acmart}
\AtBeginDocument{%
  }

\copyrightyear{2025}
\acmYear{2025}
\setcopyright{licensedothergov}\acmConference[SIGIR '25]{Proceedings of the 48th International ACM SIGIR Conference on Research and Development in Information Retrieval}{July 13--18, 2025}{Padua, Italy}
\acmBooktitle{Proceedings of the 48th International ACM SIGIR Conference on Research and Development in Information Retrieval (SIGIR '25), July 13--18, 2025, Padua, Italy}
\acmDOI{10.1145/3726302.3730302}
\acmISBN{979-8-4007-1592-1/2025/07}
\usepackage{hyperref}
\usepackage{url}
\usepackage{nicefrac}       
\usepackage{listings}
\usepackage{multirow}
\usepackage{rotating} 
\usepackage{algorithm}
\usepackage{algorithmic}
\usepackage[T1]{fontenc}

\newcommand*{\name}{LUMA}

\lstdefinestyle{promptstyle}{
    backgroundcolor=\color{teal!10},
    basicstyle=\ttfamily,
    frame=single,
    breaklines=true,
    columns=fullflexible,
    keepspaces=true,
    showstringspaces=false,
    captionpos=b,
    frameround=tttt,
}

\begin{document}

\title{LUMA: A Benchmark Dataset for Learning from Uncertain and Multimodal Data}

\author{Grigor Bezirganyan}
\email{grigor.bezirganyan@univ-amu.fr}
\affiliation{%
  \institution{Aix Marseille Univ,
  CNRS, LIS}
  \city{Marseille}
  \country{France}
}

\author{Sana Sellami}
\email{sana.sellami@univ-amu.fr}
\affiliation{%
  \institution{Aix Marseille Univ,
  CNRS, LIS}
  \city{Marseille}
  \country{France}
}

\author{Laure Berti-\'Equille}
\email{laure.berti@ird.fr}
\affiliation{%
  \institution{IRD, ESPACE-DEV}
  \city{Montpellier}
  \country{France}
}

\author{S\'ebastien Fournier}
\email{sebastien.fournier@univ-amu.fr}
\affiliation{%
  \institution{Aix Marseille Univ,
  CNRS, LIS}
  \city{Marseille}
  \country{France}
}


\begin{abstract}
Multimodal Deep Learning enhances decision-making by integrating diverse information sources, such as texts, images, audio, and videos. To develop trustworthy multimodal approaches, it is essential to understand how uncertainty impacts these models. We propose  LUMA, a unique multimodal dataset, featuring audio, image, and textual data from 50 classes, specifically designed for learning from uncertain data. It extends the well-known CIFAR 10/100 dataset with audio samples extracted from three audio corpora, and text data generated using the Gemma-7B Large Language Model (LLM). The LUMA dataset enables the controlled injection of varying types and degrees of uncertainty to  achieve and tailor specific experiments and benchmarking initiatives. LUMA is also available as a Python package including the functions for generating multiple variants of the dataset with controlling the diversity of the data, the amount of noise for each modality, and adding out-of-distribution samples. A baseline pre-trained model is also provided alongside three uncertainty quantification methods: Monte-Carlo Dropout, Deep Ensemble, and Reliable Conflictive Multi-View Learning. This comprehensive dataset and its tools are intended to promote and support the development, evaluation, and benchmarking of trustworthy and robust multimodal deep learning approaches. We anticipate that the LUMA dataset will help the research community to design more trustworthy and robust machine learning approaches for safety critical applications. The code and instructions for downloading and processing the dataset can be found at: \href{https://github.com/bezirganyan/LUMA/}{https://github.com/bezirganyan/LUMA}.

\end{abstract}




\maketitle

\section{Introduction}

In recent years, the use of Machine Learning and Deep Learning has surged across various fields, driving advancements in data analysis and decision-making. In domains such as healthcare, autonomous driving, and finance, information is distributed across multiple modalities including audio, video, text, and images. To better understand the data and improve decision-making capabilities, it is crucial for deep learning models to integrate  diverse, multimodal sources of information. Multimodal Deep Learning (MDL) addresses this need and improves the capabilities of uni-modal networks \citep{bayoudh_surveydeep_2022, Krones2024ReviewOM, xiao2020multimodal, lee2020multimodal}. 

Another important consideration for deploying deep learning models in safety critical fields is trustworthiness. Traditional deep learning models are often overconfident in their predictions \citep{abdar2021review}, which can lead to catastrophic results in areas such as healthcare or autonomous driving. Although various techniques for uncertainty quantification have been proposed to measure the level of uncertainty in data and model, this remains an open and challenging area. More research and robust benchmarks are needed to advance the field of uncertainty quantification in deep learning \citep{krishnan2020improving, nado2021uncertainty}.

In probabilistic modeling,  uncertainty is usually divided into aleatoric (data) and epistemic (model) uncertainties \citep{kiureghian_aleatoryepistemic_2009}. Aleatoric uncertainty  refers to the uncertainty in data due to inherent noise. It is impossible to reduce the amount of aleatoric uncertainty with additional data (hence, it is also often called irreducible uncertainty). Epistemic uncertainty is the uncertainty in model parameters, due to lack of data, hence, it can be reduced with additional data samples. Epistemic uncertainty is also usually high for Out-of-distribution (OOD) data, and is commonly used for OOD detection. 

Multimodal uncertainty quantification (MUQ) is a relatively new research area that adapts uncertainty quantification approaches to multimodal deep learning problems, aiming to enhance the trustworthiness of these models \citep{jung_uncertaintyestimation_2022}. Due to the unsupervised nature of uncertainty quantification, where the exact extent of uncertainty in the data and the model is unknown, analyzing and benchmarking proposed MUQ methods is challenging. Current multimodal datasets used for benchmarking state-of-the-art models in multimodal uncertainty quantification \citep{han_trustedmultiview_2021, jung_uncertaintyestimation_2022, jung_unimodalgeneralising_2023, liu_trustedmultiview_2022, Xu_Si_Guan_Zhao_Wu_Gao_2024} lack the ability to inject a controlled amount and various types of uncertainties for each modality. This limitation hinders the comprehensive benchmarking of MUQ techniques, which is essential for developing trustworthy and robust multimodal deep learning approaches.


To address this challenge, we introduce \name{} ({L}earning from {U}ncertain and {M}ultimodal d{a}ta), a multimodal dataset specifically designed for benchmarking multimodal learning algorithms on uncertain data. The dataset includes 101,000 images, 135,096 speech audio recordings, and 62,875 text passages, amounting to approximately 3 GB of data. Each modality is independently sourced, reflecting real-world conditions where data is often collected under different conditions and times. For example, in medical contexts, diagnostic data from different modalities such as radiography, MRI, and ECG/EEG are gathered asynchronously, leading to modality-specific uncertainties. The modalities are carefully aligned, ensuring that each text passage is related to the object in the corresponding image, and each audio recording is the pronunciation of the object label in the image. The provided Python toolkit allows the injection of aleatoric and epistemic 
uncertainties in a controlled and parameterized way into each  modality specifically.




To summarize, our contributions are as follows:
\begin{enumerate}
    \item We propose  \name{}\footnote{\href{https://huggingface.co/datasets/bezirganyan/LUMA}{https://huggingface.co/datasets/bezirganyan/LUMA}}, a multimodal dataset specifically designed for learning from uncertain data. It includes audio, image, and textual modalities across 50 distinct classes. We compiled the images from the CIFAR 10/100 dataset \citep{krizhevsky_learningmultiple_}, extracted, validated, and associated audio samples corresponding to the CIFAR image class labels from three diverse audio corpora~\cite{arne_mining_2016,panayotov_librispeechasr_2015,commonvoice:2020}, and generated the text modality based on the class labels using the Gemma-7B Instruct \citep{gemmateam_gemmaopen_2024} Large Language Model (LLM). We also performed additional bias analysis of the dataset. Each generated version of the dataset consists of 600 data records per class (500 for training, and 100 for testing) belonging to 42 classes, and 3,859 OOD data points, belonging to the remaining 8  classes. 
    \item We offer a Python package\footnote{\href{https://github.com/bezirganyan/LUMA}{https://github.com/bezirganyan/LUMA}} that generates dataset samples with varying levels of noise and uncertainty. The uncertainty generator can effectively increase aleatoric uncertainty in the data and epistemic uncertainty in the model. 
    \item Finally, we provide baseline models including  three  different uncertainty quantification methods (Monte-Carlo Dropout \citep{gal2016dropout}, Deep Ensemble  \citep{lakshminarayanan2017simple}, Reliable Conflictive Multi-View Learning \citep{Xu_Si_Guan_Zhao_Wu_Gao_2024}), to serve as a starting point for benchmarking. 
\end{enumerate}
\section{Limitations of current Datasets for MDL benchmarking}

In practice, we often don't know the extent of inherent uncertainties in the data or how accurately they represent the real-world data space. This often makes it hard to evaluate how well uncertainty quantification algorithms work. Moreover, deep learning algorithms may behave differently under different amount of uncertainties (i.e.,  the robustness to noise may vary). Thus, it may be beneficial to inject additional amount of noise in the data, and observe the change in uncertainty metrics and the performance of the models. Since approaches to quantify different types of uncertainty vary, it is beneficial to have options for injecting various types of uncertainties. 

Several datasets are used in multimodal uncertainty quantification settings. A notable line of work  \citep{han_trustedmultiview_2021, jung_uncertaintyestimation_2022, jung_unimodalgeneralising_2023} has employed datasets such as HandWritten\footnote{https://archive.ics.uci.edu/ml/datasets/Multiple+Features}, CUB\footnote{http://www.vision.caltech.edu/visipedia/CUB-200.html}, Scene15\footnote{https://serre-lab.clps.brown.edu/resource/hmdb-a-largehuman-motion-database}, and Caltech101~\footnote{https://data.caltech.edu/records/mzrjq-6wc02}. 
These datasets typically extract different features from unimodal sources to create a multi-view setup. While they have been instrumental, they primarily repurpose unimodal data for multimodal tasks, underscoring the need for more comprehensive and inherently multimodal datasets to better evaluate uncertainty in deep learning models.

Furthermore, the current  approaches that introduce uncertainty in the data \citep{han_trustedmultiview_2021, jung_uncertaintyestimation_2022, jung_unimodalgeneralising_2023} add Gaussian noise to the views or the extracted features. While Gaussian noise does increase uncertainty, it does not accurately reflect the noise that can be found in real-world datasets and this process lacks fine-grained control over the type of uncertainty being injected.

Additionally, how different modalities' uncertainties interact significantly impacts the overall multimodal uncertainty. When both modalities encode redundant information, the total uncertainty might not decrease. Conversely, conflicting information can lead to increased uncertainty, while complementary information can reduce it. A deeper understanding of these phenomena is crucial. Fine-grained control over individual modalities' uncertainties opens the way for more theoretical research based on empirical observations.

To better understand and analyze uncertain multimodal data, as well as to debug and benchmark uncertainty quantification techniques in the multimodal learning context, we propose a dataset accompanied by an uncertainty generator package. This package includes various techniques for injecting uncertainty, such as controlling data diversity, adding different types of real-world noise, randomly switching labels to their closest class, and injecting out-of-distribution (OOD) data.
\section{\name{} Dataset}

In this section, we introduce  \name{}, a dataset composed of an extensible list of modalities including image, audio, and text modalities, collected from various sources. 

\begin{figure}[h]
    \centering
    \includegraphics[width=1\linewidth]{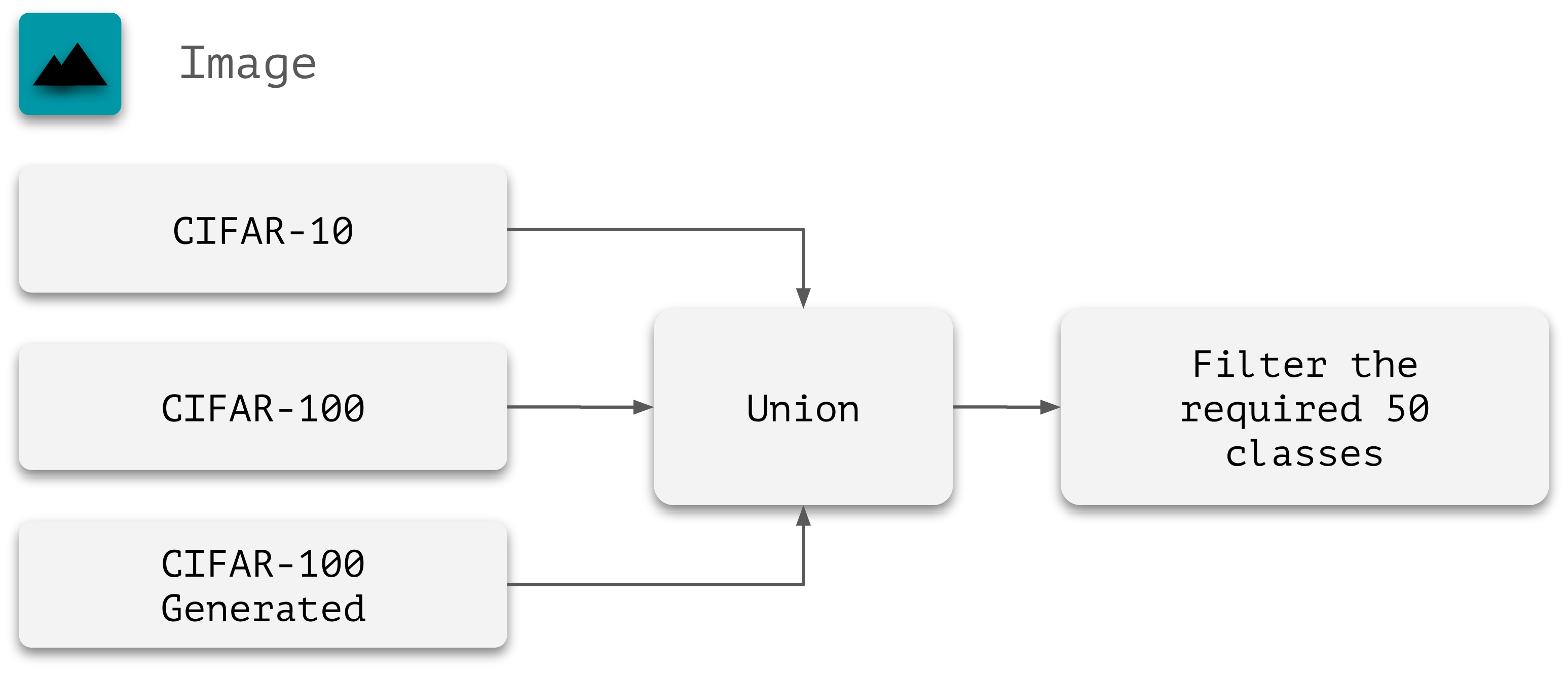}
    \caption{Image collection pipeline}
    \label{fig:image-pipeline}
    \Description{Figure 1: Fully described in the text.}
\end{figure}

\subsection{Image modality}
\label{sec:image_modality}
For the image modality, our priority was to choose a relatively simple yet well-known dataset, where we could have the option to manually increase the degree of uncertainty. For that purpose, we chose CIFAR-100 and CIFAR-10 \citep{krizhevsky_learningmultiple_} datasets since they are well-known datasets of small 32x32 images, with lots of baseline models. 42 classes were chosen so that after aligning with the other modalities, we would have at least 600 samples in each class per modality. The threshold of 600 classes was selected based on the number of images per class in the CIFAR-100 dataset. We took another 8 classes, which had less than 600 samples after aligning with other modalities, as OOD samples. In total, we took 25,200 images as train/test data, and 3,859 images as OOD data (see the image collection pipeline in Figure \ref{fig:image-pipeline}).

Aside from the main dataset, as described in Section \ref{subsub:diversity}, another priority was to understand the behaviors of models under different levels of data diversity. To achieve this, we decided to sample 600 data points with different level of diversity from the bigger set CIFAR-10/100. However, in CIFAR-100 dataset, there are no more than 600 samples per class. We alleviated this issue with including images generated with EDM Diffusion-based generative model\footnote{Retrieved generated samples from https://github.com/wzekai99/DM-Improves-AT}\citep{Karras2022edm}. We chose EDM-generated images, since the generated samples were already available, and \citet{zheng2024toward} showed that augmenting CIFAR-10 data with EDM-generated samples significantly improves the classification accuracy. 

\subsection{Audio modality}
\label{sec:audio_modality}
\begin{figure}[h]
    \centering
    \includegraphics[width=1\linewidth]{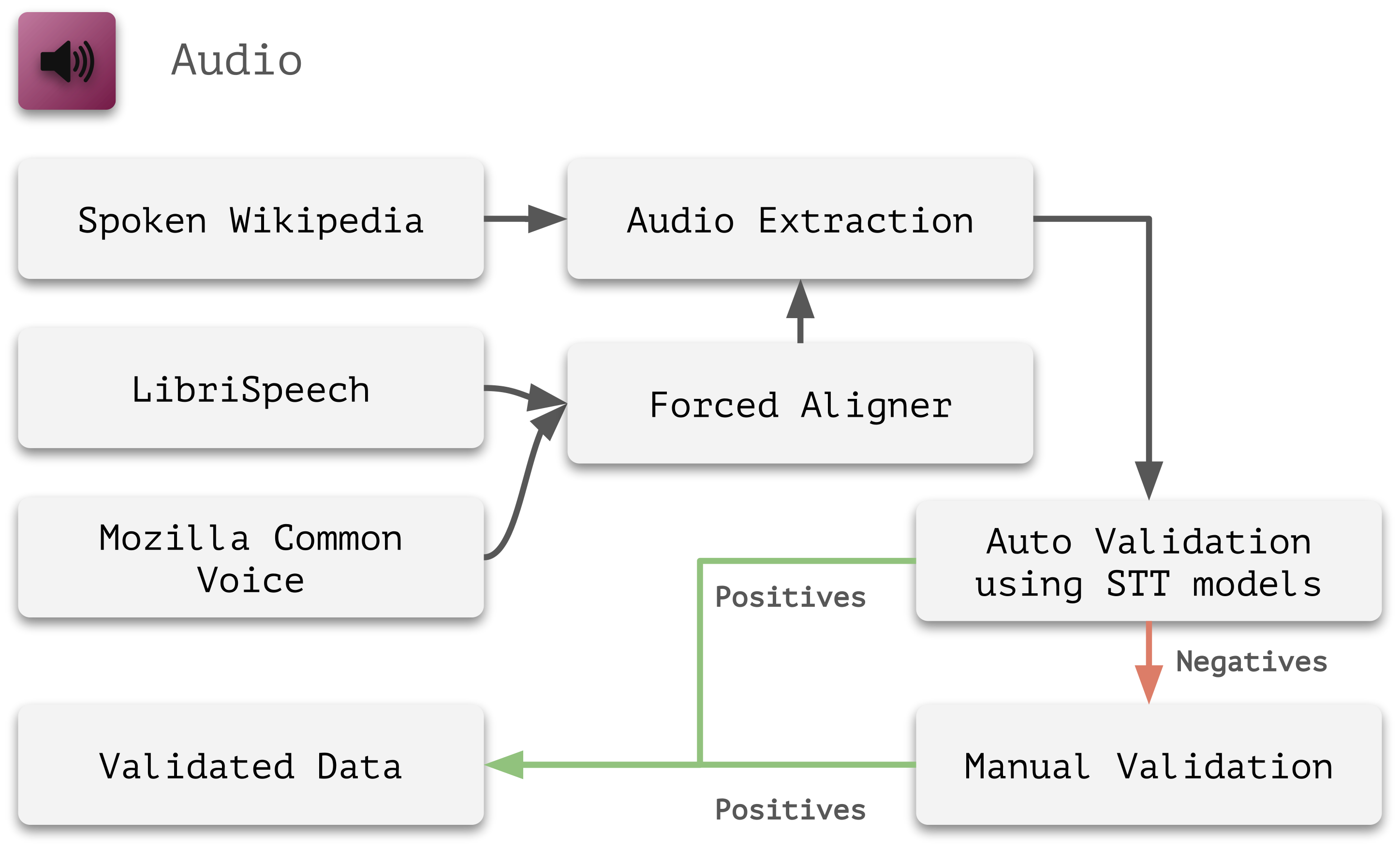}
    \caption{Audio collection, extraction and validation pipeline}
    \label{fig:audio-extraction-pipeline}
    \Description{Figure 2: Fully described in the text.}
\end{figure}

For audio modality, the diversity of accent in the pronunciation was an important factor to consider, and we collected samples, where different people would pronounce the corresponding class label of CIFAR 10/100 images. For this task, we used three audio/text parallel corpora and extracted the desired audio segments. More specifically, we used The Spoken Wikipedia \citep{arne_mining_2016}, LibriSpeech \citep{panayotov_librispeechasr_2015}, and Mozilla Common Voice \citep{commonvoice:2020} corpora. The audio collection pipeline is shown in Figure \ref{fig:audio-extraction-pipeline}.

The Spoken Wikipedia is a collection of hundreds of hours of phoneme-level aligned audio, where volunteer readers are reading various Wikipedia articles. We used these alignments to extract all the instances of audio segments that pronounced one of the CIFAR-10/100 classes.

The LibriSpeech dataset is a corpus of 1,000 hours of English speech, derived from audiobooks from the LibriVox\footnote{https://librivox.org/} project, which is a collection of public domain audiobooks. Unfortunately, LibriSpeech doesn't provide word-level alignment, hence, we used force-aligned alignments\footnote{https://github.com/CorentinJ/librispeech-alignments} generated with the Montreal Forced Aligner \citep{mcauliffe17_interspeech}. Similarly to The Spoken Wikipedia, we looked up the CIFAR-10/100 labels in forced aligned textual data, and extracted the corresponding audio segments. 

The Mozilla Common Voice corpora, is a crowdsourced open-source collection of voices by volunteer contributors from around the world. Like LibriSpeech, Mozilla Common Voice also doesn't provide word-level alignments, hence, we again used forced aligned alignments\footnote{https://github.com/JRMeyer/common-voice-forced-alignments}, and extracted the relevant audio samples. 

73 additional recordings of pronunciations belonging to 4 classes ("roses", "telephone", "whale", "wolf") were voluntarily contributed by our colleagues, which were anonymized, trimmed, and added to the dataset.

From these corpora, we used the following rule to extract the samples. First, we extended our class label set with a superset that also contains the plural forms of the words (i.e., for the audio track  ``horse'', the audio track ``horses'' was added to the set), then we iterated over all aligned transcripts, and for any word included in the formed set, we extracted the corresponding audio sample. We considered the plural forms, since we believe that an extra "s" or "es" does not change the pronunciation of the words much. We did not consider the plural forms if it requires audible changes to the word root (i.e., mouse - mice).
Since most of the audio data is collected from forced alignments, it is possible to have misaligned audio segments, which could introduce additional noise to the dataset. Moreover, since part of the audio samples are from voluntary contributions, there can be very noisy samples, which are hard to interpret, or audio samples with a strong accent, which again can be hard to interpret. To remove such extreme cases in audio samples, we performed an automatic validation of the samples. Then, we filtered out the false negatives with manual validation for the negative predictions. 

The automatic validation was achieved with the OpenAI's Whisper Large V3 model \citep{radford2022whisper} for audio transcription, and transcribed the extracted audio samples. If the transcription corresponded to the class label (or its plural form), then we considered the sample as valid. Otherwise, the sample was sent for manual validation. Because of the huge output space of the Whisper Large V3 model, the probability of false positives is quite low, so we did not perform a manual validation for positive predictions. To summarize,  we validated 130,069 out of 178,123 data samples with automatic validation, and  we performed a manual validation for the remaining samples. 


For manual validation, we decided to check only the classes, which did not have more than 800 samples (to be able to sample 600 samples with different degrees of diversity, as described in Section \ref{subsub:diversity}). Hence, we filtered 8,372 samples, and scheduled them for manual labeling. We opted for Label Studio \citep{LabelStudio} to build the labeling interface. 
 The interface provided the audio sample, with the prompt "Is the audio saying the word below? (An extra 's' or 'es' in the pronunciation is okay.)" and answer options of "Yes" or "No". We asked our colleagues (M.Sc. and PhD students, and professors) with advanced to fluent English knowledge to annotate the samples.  

\begin{figure}
    \centering
    \includegraphics[width=1\linewidth]{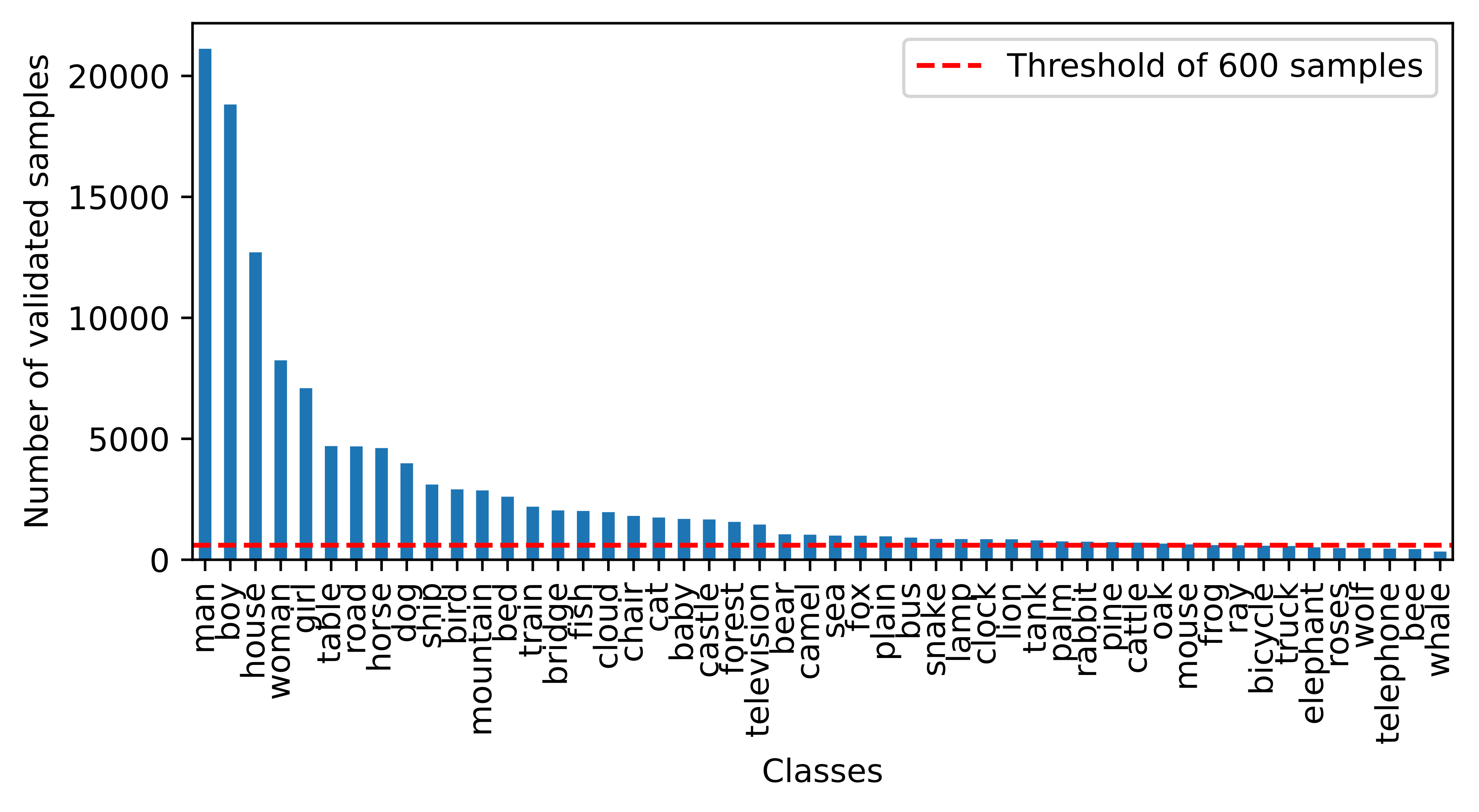}
    \caption{Number of validated audio samples per class. We will include the classes with higher than 600 samples as in-distribution data, and others as out-of-distribution data.}
    \label{fig:audio-validated-data-dsitribution}
    \Description{In the validated audio samples there are classes with much higher than 600 samples, and classes that have less than 600 validated samples.}
\end{figure}

In total, we collected 2 annotations per sample, from 17 annotators. We got 71.61\% annotation agreement and accepted 5,027 samples, where both annotators confirmed the validity of the sample. Samples with conflicting annotations were rejected. Hence, we took the 42 classes that had more than 600 validated samples (automatically and manually) as training / test data, and the remaining 8 classes as OOD data. In total, the auto-validated and manually validated audio samples combined, \name{} has 135,096 audio samples. 
The final distribution of audio data across classes can be seen in Figure \ref{fig:audio-validated-data-dsitribution}.

\subsection{Text Modality}
\label{sec:text_modality}

For text modality,  the main constraint was that the text segments had to talk about the label of the images. For that, we decided to employ a generative model, and generate text segments about the class label. We utilized Google's Gemma-7B Instruct model \citep{gemmateam_gemmaopen_2024} to generate more than 1,200 texts samples per class, using 13 different prompts. The 13 prompts are as follows:
\begin{lstlisting}[style=promptstyle]
"You are talking with your friend about some topic. Use the word <word> in a sentence with your friend. Use casual language. Tone: Casual / Conversational, length: short",
"You are the prime minister of the United Kingdom. During a press conference you are asked a question about <word>. Give a sentence from that press conference mentioning the word <word>. Tone: Formal, length: medium",
"You are explaining a five year old child what the word <word> means. Use very simple and explanatory language, so the kid will understand the meaning of the word <word>. Tone: Casual, complexity: simple",
"Imagine you are writing a science fiction book. Write a conversation from that book mentioning the <word>.",
"You are the editor in a mainstream journal. Write a sentence from a news article about a <word> in your journal that mentions the word <word>.",
"You are a teenager writing a post in Facebook about <word>. Write the post about the experience you had with the <word>.",
"You are playing a word describing game with your friend. The word is <word>, and you shall describe it without mentioning the word itself, so your friend will guess it. Explain it to him clearly in a simple language.",
"Think of something else that shares similar characteristics or functions with the <word>. Draw comparisons or use analogies between that other word and the <word>.",
"Place the word <word> within historical context. How would you describe it in relation to its origins, evolution, or significant historical events? Be creative in your description.",
"Consider how the word '<word>' is depicted or referenced in popular culture, literature, or media. Describe it by referencing these cultural elements.",
"Pretend you are a character who sees a <word> for the first time in your life. Describe it from the character's perspective, considering their background, personality, and knowledge.",
"Write a 4 line small poem about the word <word>. Be creative, and use casual tone for the poem.",
"You are a musician composing a song inspired by <word>. Write the lyrics to the song, capturing the mood, emotions, and imagery associated with <word>. Use rhythm and melody to convey the essence of <word> in your music."
\end{lstlisting}
The \texttt{<word>} was replaced with the class labels.
Gemma-7B Instruct was chosen, since according to their technical report \citep{gemmateam_gemmaopen_2024},  it outperforms other open LLMs with similar size, in 11 out of 18 tasks. Moreover, in our experiments, it provided responses that we found more useful for our intended applications than those from Mistral-7B \citep{jiang2023mistral}.

\begin{figure}
    \centering
    \includegraphics[width=1\linewidth]{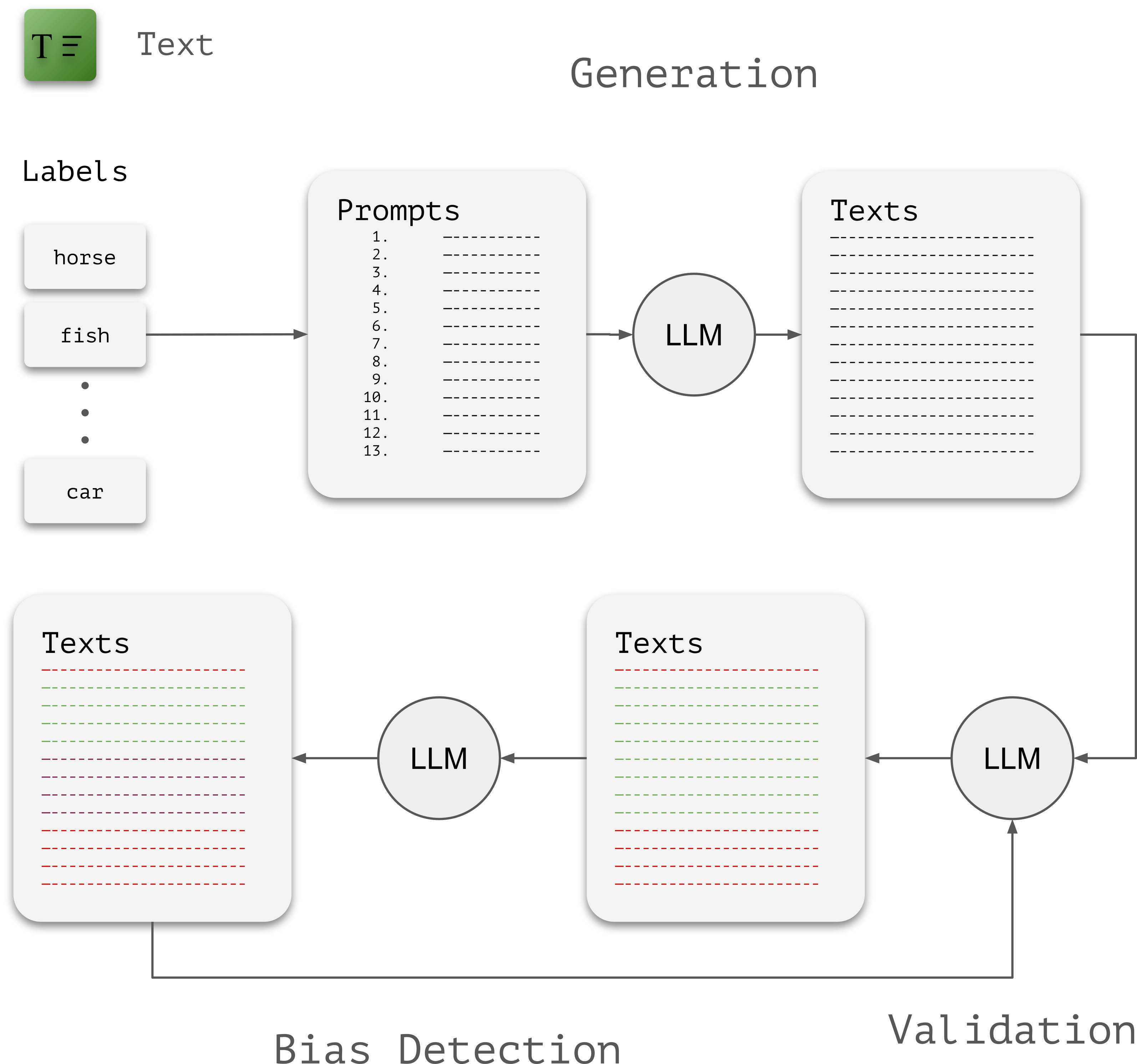}
    \caption{Text generation and validation pipeline}
    \label{fig:text-pipeline}
    \Description{Figure 4: Fully described in the text.}
\end{figure}

\begin{figure}
    \centering
    \includegraphics[width=0.8\linewidth]{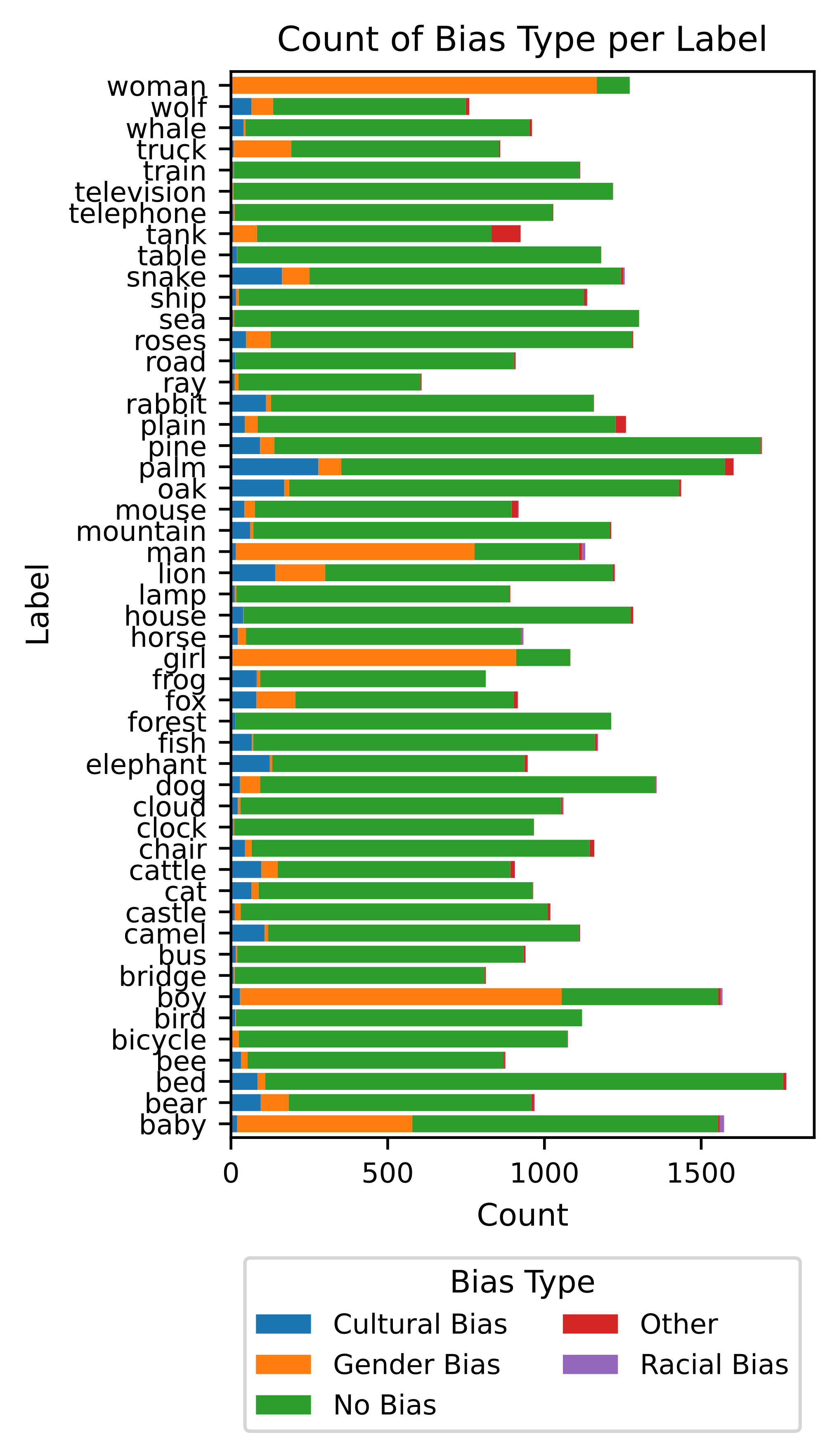}
    \caption{The amount of texts with different biases according to Gemma-7B Instruct model.}
    \label{fig:bias_count}
        \Description{Woman, Man, Girl and Boy classes have the highest amount of biased texts, while other classes have very low amount of biases.}
\end{figure}

To validate that the generated texts accurately represent the labels, we masked all label occurrences in the text and fed the masked text back into the Gemma-7B Instruct model, asking it to classify the text into one of the labels. Based on the prediction of the model, if the prediction matched the ground truth label, we accepted the sample as validated. In total, we accepted 55,953 text samples. 

After manually analyzing some of the generated texts, we noticed that there were samples with offensive biases and stereotypes, such as:
\begin{lstlisting}[style=promptstyle]
A woman is a grown-up person who has a soft, nurturing personality. She usually takes care of her family and friends, and sometimes works outside the home. Women are strong and smart, they can do many things that men can do.

A tool in a toolbox is an efficient and valuable asset that aids in various tasks. Similarly, a man is also a valuable asset to any group or society. Just like a tool in a toolbox, a man's capabilities are tailored to fulfill different roles and functions, making him an essential component of any endeavor.
\end{lstlisting} 
Particularly, we noticed a lots of gender bias for classes "man", "woman", "boy" and "girl". To find the proportion of the biased data, we asked the Gemma model to find out if the given text contains gender, racial, religious, or cultural biases. We found out, that indeed, the aforementioned 4 classes have a huge amount of gender bias (See Figure \ref{fig:bias_count}). Our hypothesis is, that describing a man or woman in an unbiased way is a challenging task for LLM models (as well as for humans), which are trained on unbalanced data \citep{kotek2023gender}. Although the model identified a high level of bias in these classes, indicating a potential for LLM-based bias detection, we are unable to assess how accurate or consistent these detections truly are.

\begin{figure}[h]
    \centering
    \includegraphics[width=0.8\linewidth]{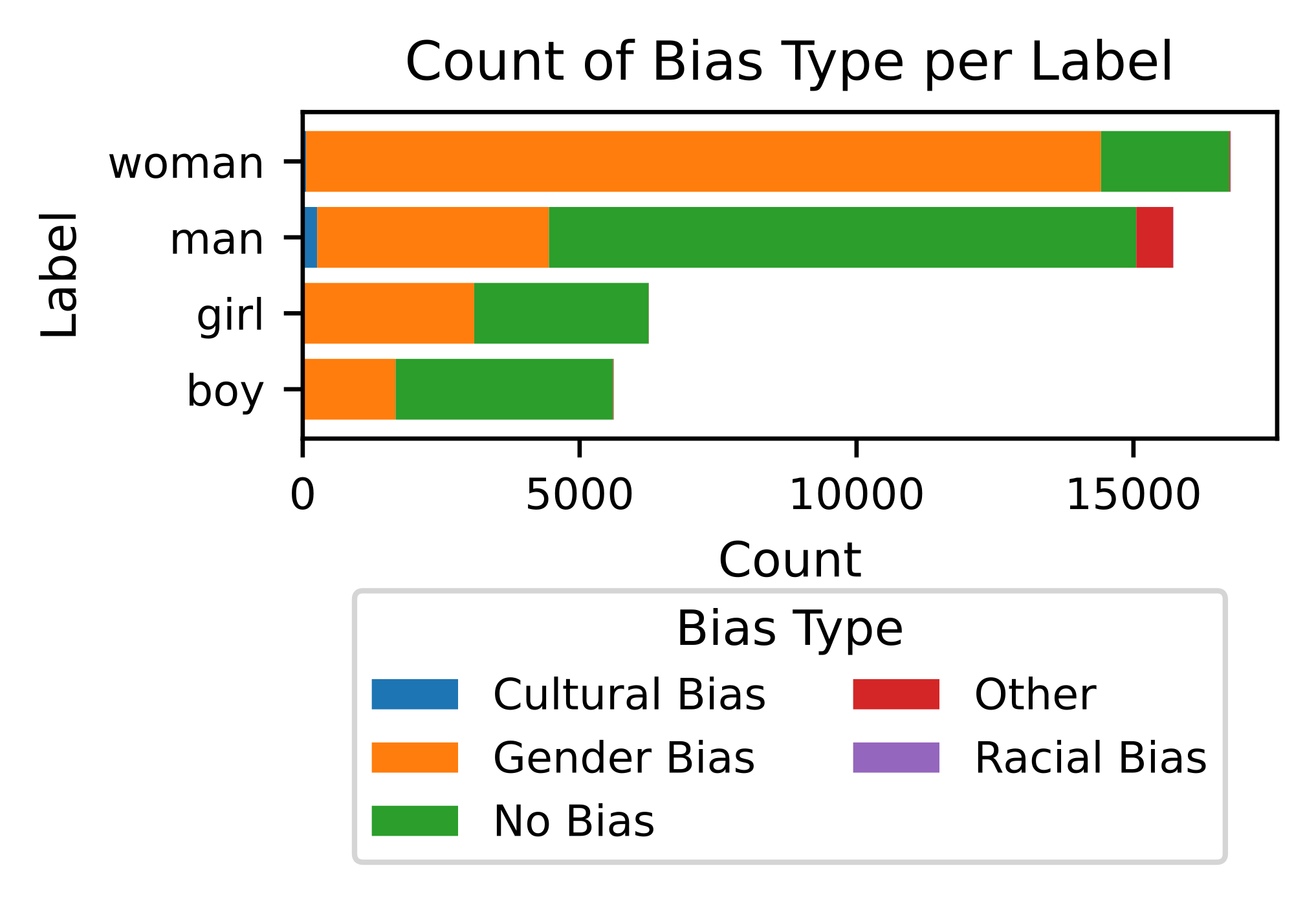}
    \caption{The amount of texts with different biases according to Gemma-7B Instruct model, after reconstructing the prompts for these 4 labels. Although there is still high amount of bias, we can filter them out and still have enough unbiased texts (more than 800 text passages per class).}
    \Description{Regenerated texts have enough unbiased samples to be included in the dataset.}
    \label{fig:bias_count_after_unbiasing}
\end{figure}
To reduce the biases for the labels of \texttt{man}, \texttt{woman}, \texttt{boy} and \texttt{girl}, we reconstructed the prompts to explicitly provide topics and keywords with occupations, which will minimize the bias. The prompt used is the following: 
\newline
\begin{lstlisting}[style=promptstyle]
Write two sentences with topic: <topic>, and keywords: <keyword1>, <keyword2>.
\end{lstlisting}
where \texttt{<topic>} was replaced with one of the following words:  \texttt{`factual', `fiction', `history', `books', `movies',} \texttt{}{ `philosophical'}, and \texttt{<keyword1>} with one of the following words: \texttt{`man', `woman', `boy', `girl'}. \texttt{<keyword2>} was treated differently depending on the label: For the term `man', it was replaced with one of the following words: 
\begin{lstlisting}[style=promptstyle]
actor, king, scientist, doctor, wizard, duke, lord, governor, prime minister, father, sorcerer, waiter, chess, director, producer, uncle, singer 
\end{lstlisting}
For the term \texttt{woman}, it was replaced with one of the following words:
\begin{lstlisting}[style=promptstyle]
actress, queen, scientist, doctor, witch, duchess, lady, governor, prime minister, mother, sorcerer, waitress, chess, director, producer, aunt, singer 
\end{lstlisting}
For the term \texttt{boy}, it was replaced with one of the following words:
\begin{lstlisting}[style=promptstyle]
kid, actor, prince, son, nephew, pupil, student, singer
\end{lstlisting}
And for the term \texttt{girl}, it was replaced with one of the following words:
\begin{lstlisting}[style=promptstyle]
kid, actress, princess, daughter, niece, pupil, student, singer
\end{lstlisting}
We then performed another round of bias detection using the Gemma model. While we found that a significant amount of bias still exists, we identified enough unbiased texts (according to the Gemma model) to include in the LUMA dataset. The number of biased and unbiased texts after re-generating the data for these 4 classes can be found in Figure \ref{fig:bias_count_after_unbiasing}.

Since textual data was generated using an LLM, we recognize that the dataset may contain factual inaccuracies, or biases, but our aim is to offer a benchmark to study uncertainty quantification in multimodal classification settings. \name{} shall not be used as a source of knowledge or information.

\subsection{Dataset compilation}

Based on the collected samples from the 3 modalities (image, text, audio), we first compiled a clean version of the dataset with minimal uncertainty. Our goal was to then provide tools that allow uncertainty to be introduced on demand. We prioritized flexibility, offering multiple options for controlling uncertainty through adjustable parameters such as data diversity, sample-level noise, label corruption, and the injection of out-of-distribution (OOD) samples.

\label{subsub:diversity}

\begin{figure}
    \centering
    \includegraphics[width=1\linewidth]{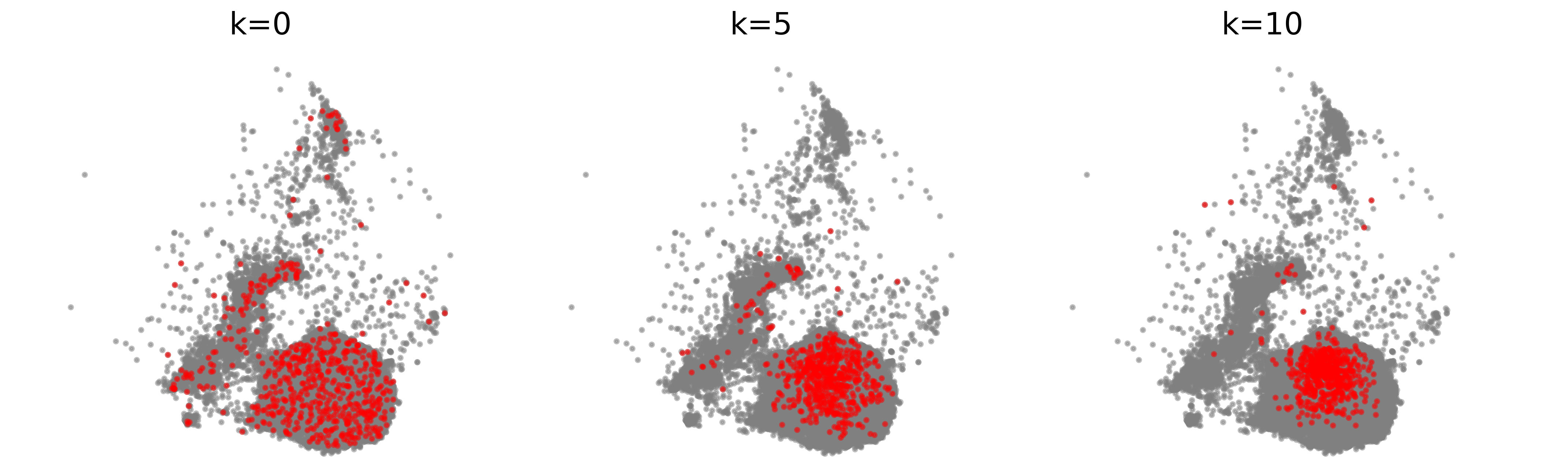}
    \caption{t-SNE \citep{tsne} visualization of audio data points from class "man" (in gray), and sampled points with different diversity parameter $k$. The higher is the value of $k$, the more concentrated (less diverse) the points are.}
    \label{fig:class-diversity}
    \Description{Figure 7: Fully described in the caption and text.}
\end{figure}

\subsubsection{Data Diversity}
\begin{figure}
    \centering
    \includegraphics[width=1\linewidth]{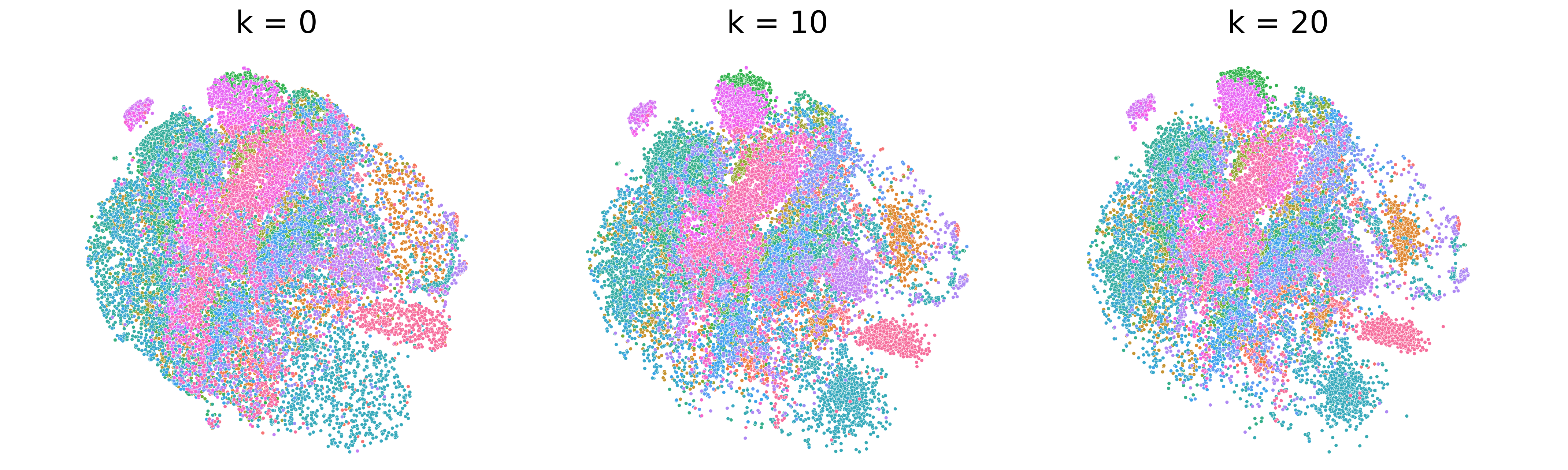}
    \caption{t-SNE \citep{tsne} visualization of audio data points for all classes, sampled with different diversity parameter $k$. With higher $k$ we have more concentrated samples, and more separation between classes. The diversity can similarly be controlled for the other modalities.}
    \label{fig:total-diversity}
    \Description{Figure 8: Fully described in the caption and text.}
\end{figure}

With a fixed number of data points, increasing the diversity of data enhances the information passed to the model; thereby, it shall reduce the epistemic uncertainty. Conversely, when samples are concentrated at a single point in the latent space, they encode less information, which shall lead to greater epistemic uncertainty in areas where data is scarce (Figure \ref{fig:total-diversity}). Hence, controlling the diversity of the data allows us to study the behavior of epistemic uncertainties under varying amounts of information.

To control the diversity, we extract deep features from each modality (Wav2Vec \citep{baevski2020wav2vec} for audio, BERT \citep{devlin2019bert} for text and VGG-11 \citep{simonyan2015very_deep} for images), and compute the inverse distance of each sample to the center (mean vector) of its class, raised to the power of $k$:
\begin{equation}
    \label{eq:distance}
    D_i=\frac{1}{\left\|F_i-\frac{1}{|C|} \sum_{j \in C} F_j\right\|_2^k}, \ \ i \in C, 
\end{equation}
where $F$ represents the deep feature vectors extracted from the samples, $C$ is the set of data indices belonging to the class, and $|\cdot|$ measures the cardinality of the set. Then, having the inverse distances, we sample points from the categorical distribution \mbox{$x_n \sim \text{Cat}(D)$}. In Eq. \ref{eq:distance}, $k$ is the variable controlling the diversity. If $k=0$ the sampling is uniform. The bigger $k$, the higher probability of selection will be applied to the samples closer to the center.  

Having sufficient samples in image and text modalities, our bottleneck was the number of samples in the audio modality. Since in the 42 in-distribution classes, around 70\% have more than 900 audio samples, we considered this enough for diversity control. 


\subsubsection{Sample Noise}
We want to have an option to inject a controlled amount of noise into the data. This may  reduce the information in each data sample and increase the classification difficulty. With our hypothesis, this may affect the aleatoric uncertainty degrees. This type of noise can also be very beneficial for estimating the  model's robustness to noise. We apply different types of noise to each modality.

For \textbf{audio modality}, we added background noise (Human non-speech sounds, urban noises, animal sounds, natural soundscapes, etc.) from the ESC-50 dataset \citep{piczak2015esc} to each sample, using the  \texttt{audio-} \texttt{mentations}\footnote{https://github.com/iver56/audiomentations} library. The amount of the minimum and maximum signal-to-noise ratio, as well as the proportion of the noisy data, is set as a hyper-parameter.  

For \textbf{text modality}, we utilize the \texttt{nlpaug} \citep{ma2019nlpaug} library, to add different types of noise. The user has the option to choose a subset of noise types from: 1) Keyboard noise that simulates keyboard distance error; 2) OCR noise that simulates OCR engine noise; 3) Random character noise to insert, substitute, or delete random characters; 4) Antonym noise to swap random words with their antonyms; 5) Random word noise to insert, substitute, or delete random words; 6) Spelling noise to add spelling mistakes according to the spellingling mistake dictionary; and 7) Back-translation noise to translate the text to another language, and then translate back to English.
The parameters of these noise types can be specified by the user, and are transferred to the \texttt{nlpaug} library for adding the specific type of noise. 

For \textbf{image modality}, we added different types of noise suggested and implemented by \citet{hendrycks2018benchmarking}. 15 perturbations are included such as: adding Gaussian noise, shot noise, impulse noise, defocus blur, frosted glass blur, motion blur, zoom blur, snow, frost, fog, changing the brightness, contrast, elasticity, pixelating, and JPEG compressing. Additionally, common transformations such as cropping or skewing are not part of the default set but can be easily applied by the user outside the framework if needed.

\subsubsection{Label Noise}
Aleatoric uncertainty can also be introduced by injecting label noise into the dataset, i.e., by randomly altering the labels of certain samples. Since this type of uncertainty arises from inherent noise in the data itself, it cannot be mitigated by increasing the dataset size and thus directly contributes to the model's aleatoric uncertainty.

To inject label noise in a more structured manner, we randomly select a subset of samples based on a user-defined probability. For each selected sample, we compute its deep feature representation using modality-specific encoders: Wav2Vec \citep{baevski2020wav2vec} for audio, BERT \citep{devlin2019bert} for text, and VGG-11 \citep{simonyan2015very_deep} for images. We then measure the sample’s average distance to the five nearest samples from each class. The sample is reassigned the label of the class with the lowest mean distance, ensuring the new label is still semantically or perceptually close to the original, thereby creating realistic label noise.

\subsubsection{OOD Injection}
Ideally, the models shall be uncertain on data points from unknown distribution (i.e., distribution they haven't been trained on). In the literature, often the OOD samples are taken from another dataset, which can simplify the problem, because  such samples are far from the training data. For this matter, we kept a separate set of samples from the same dataset, but belonging to classes that are not present in the training data, as OOD samples.  

\section{Baseline models with Uncertainty Quantification} 

\begin{table*}[ht!]
    \centering
    \caption{Results for UQ with baseline models. The absolute values are reported for clean dataset, and changes in percentages relative to clean dataset are reported for the noisy versions of \name{}  dataset. }

    \begin{tabular}{rccrrrrrrr}
        \toprule
        \multirow{2}{*}{Method} & \multicolumn{2}{c}{Clean Dataset} & \multicolumn{2}{c}{Reduced Diversity} & \multicolumn{2}{c}{Increased Label Noise} & \multicolumn{2}{c}{Increased Sample Noise} \\
        \cmidrule(lr){2-3} \cmidrule(lr){4-5} \cmidrule(lr){6-7} \cmidrule(lr){8-9}
        & Ale. & Epi. & Ale. & Epi. & Ale. & Epi. & Ale. & Epi. \\
        \midrule
        MCD Image  & 1.00 & 1.03 & -15.73\% & -11.66\% & +59.20\% & +54.51\% & +4.44\% & +2.18\%  \\
        MCD Audio  & 0.52 & 0.70 & -5.54\% & +2.16\% & +96.63\% & +54.49\% & +23.12\% & +14.40\%  \\
        MCD Text   & 0.37 & 1.01 & -3.91\% & -2.62\% & +93.59\% & +2.41\% & +64.96\% & -2.03\%  \\
        MCD Multi. & 0.26 & 0.78 & -8.52\% & -1.21\% & +122.44\% & +11.60\% & +59.14\% & +9.89\%  \\
        DE Image   & 1.45 & 1.40 & -37.49\% & -8.54\% & -7.43\% & +0.24\% & -18.46\% & -3.22\%  \\
        DE Audio   & 0.56 & 0.99 & -27.39\% & -3.34\% & \textbf{+156.40\%} & +50.43\% & \textbf{+70.26\%} & +34.41\%  \\
        DE Text    & 0.42 & 1.01 & +5.02\% & -6.15\% & +81.26\% & -0.51\% & +62.24\% & -7.11\%  \\
        DE Multi.  & 0.31 & 0.82 & -22.80\% & -3.40\% & +115.15\% & +20.62\% & +45.97\% & +5.54\%  \\
        RCML Multi.& 1.99 & 0.43 & \textbf{+8.34\%} & \textbf{+16.16\%} & +64.72\% & \textbf{+106.16\%} & +36.19\% & \textbf{+58.21\%}  \\
        \bottomrule
    \end{tabular}
    \label{tab:uq_methods}
\end{table*}

\begin{table*}[ht]
    \centering
    \caption{OOD Detection AUC Values for Different Methods}
    \begin{tabular}{rccccccccc}
        \toprule
        \multirow{2}{*}{Method} & \multicolumn{4}{c}{MCD} & \multicolumn{4}{c}{DE} & \multirow{2}{*}{RCML Multi.} \\
        \cmidrule(lr){2-5} \cmidrule(lr){6-9}
         & Image & Audio & Text & Multi. & Image & Audio & Text & Multi. & \\
        \midrule
         AUC & 0.54 & 0.47 & 0.53 & 0.50 & 0.54 & 0.49 & 0.54 & 0.50 & \textbf{0.91} \\
        \bottomrule
    \end{tabular}
    \label{tab:ood_auc}
\end{table*}

\begin{figure*}
    \centering
\includegraphics[width=1\linewidth]{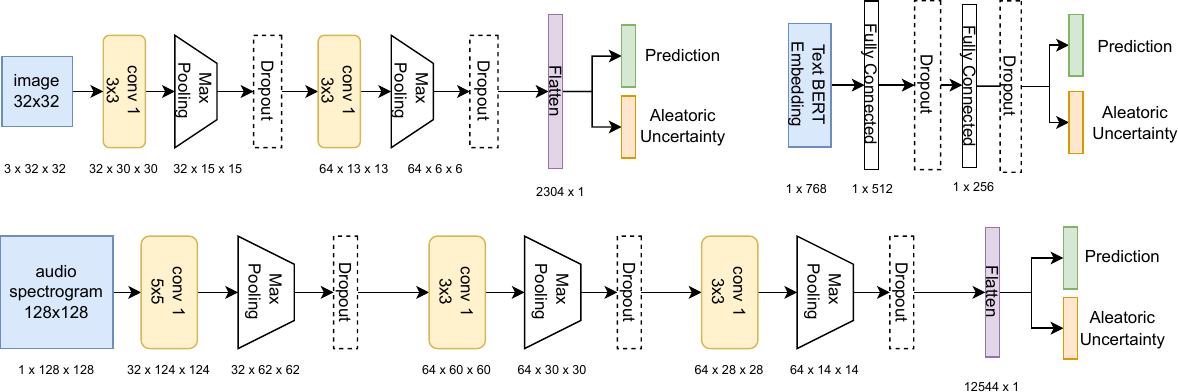}
    \caption{Network architectures used for image, audio and text modalities.}\label{fig:classifiers}
    \Description{Figure 10: Fully described in text.}
\end{figure*}

We develop baseline models with three different uncertainty quantification algorithms, to serve as a starting point for other research and benchmarking initiatives. For the sake of simplicity, we choose late fusion approaches, where we have classification networks for each modality, and then fuse their decisions by simply averaging the output logits.  These baselines were selected to instantiate unimodal and multimodal architectures, which can be trained on the dataset and are not intended to serve as a comprehensive benchmark, nor did we endeavor to achieve the best possible performance. The architectures of the baseline models are depicted in Figure \ref{fig:classifiers}.

For the image modality, we used a simple convolutional neural network. For the audio modality, we extracted 128x128 mel-spectrograms from padded audio samples, and used a convolutional network for classification. For the text modality, we extracted the BERT \citep{bert_devlin_2019} embeddings for each token, and averaged them out, so that we have one embedding per text passage. Then, we passed the embedding through a simple feed-forward neural network to get the predictions. As depicted in  Figure \ref{fig:classifiers}, each model includes two output heads: one for the prediction and the other for aleatoric uncertainty, following the methodology outlined by \citet{valdenegro2022deeper}.
Then, to combine the aforementioned unimodal networks into a multimodal architecture, we adopted the late fusion approach. In the Monte Carlo Dropout and Deep Ensemble methods, we obtained the multimodal prediction by averaging the logits from the final layers of the classifiers. For the Reliable Conflictive Multi-View Learning (RCML)~\cite{Xu_Si_Guan_Zhao_Wu_Gao_2024}, we modified the output of the last layer in each network to produce evidence, as described in \citep{Xu_Si_Guan_Zhao_Wu_Gao_2024}, and followed their methodology for combining the evidence.

 The dropout probability is 0.3, with the deep ensemble comprising 10 networks. Networks were trained for up to 300 epochs, with early stopping after 10 epochs of no validation loss improvement. 

\begin{table*}[]
    \centering
    \caption{  Classification accuracies for the clean dataset and variations in accuracy under different types of noise.
}
    \begin{tabular}{lcccc}
        \toprule
        \multirow{2}{*}{Model} & \multicolumn{1}{c}{Clean Dataset} & \multicolumn{1}{c}{Reduced Diversity} & \multicolumn{1}{c}{Increased Label Noise} & \multicolumn{1}{c}{Increased Sample Noise}  \\
        \cmidrule(lr){2-2} \cmidrule(lr){3-5}
         & \multicolumn{1}{c}{Accuracy} & \multicolumn{3}{c}{Difference in accuracy from the clean dataset} \\
        \midrule
 MCD Image &0.335 &+0.058 &-0.306 &+0.019\\
 MCD Audio &0.867 &-0.025 &-0.784 &-0.155\\
  MCD Text &0.965 &-0.027 &-0.864 &-0.144\\
MCD Multi. &0.991 &-0.010 &-0.874 &-0.063\\
  DE Image &0.387 &+0.066 &-0.166 &+0.019\\
  DE Audio &0.912 &-0.003 &-0.809 &-0.149\\
   DE Text &0.973 &-0.023 &-0.864 &-0.125\\
 DE Multi. &0.996 &-0.006 &-0.849 &-0.042\\
RCML Multi &0.973 &-0.128 &-0.833 &-0.148\\
        \bottomrule
    \end{tabular}
    \label{tab:accuracy_results}
\end{table*}

\subsection{Uncertainty Metrics}

For uncertainty quantification, we implemented 3 approaches: Monte Carlo Dropout (MCD) \citep{gal2016dropout}, Deep Ensemble (DE) \citep{lakshminarayanan2017simple}, Reliable Conflictive Multi-View Learning (RCML) \citep{Xu_Si_Guan_Zhao_Wu_Gao_2024}. 
In Monte Carlo Dropout and Deep Ensembles, we use the aleatoric entropy and the epistemic entropy as uncertainty measures, and follow \citet{valdenegro2022deeper} for disentangling the aleatoric and epistemic uncertainties.  
$H_{\text {Ale }}(y \mid \mathbf{x})=\operatorname{entropy}\left(p_{\text {Ale }}(y \mid \mathbf{x})\right)$ and 
$H_{\mathrm{Epi}}(y \mid \mathbf{x})=\operatorname{entropy}\left(p_{E p i}(y \mid \mathbf{x})\right)$, 
where $p_{Epi}$ and $p_{Ale}$ are the probabilities obtained according to \citep{valdenegro2022deeper}. For RCML~\cite{Xu_Si_Guan_Zhao_Wu_Gao_2024}, we measure the aleatoric uncertainty with the expected entropy:
\begin{equation}
\mathbb{E}_{p(\boldsymbol{\pi} \mid \mathbf{x}, \hat{\boldsymbol{\theta}})}[H[P(y \mid \boldsymbol{\pi})]]=-\sum_{k=1}^K \frac{\alpha_k}{\alpha_0}\left(\psi\left(\alpha_k+1\right)-\psi\left(\alpha_0+1\right)\right),
\end{equation} where $\alpha_k$ is the $k$-th concentration parameter of the Dirichlet distribution, and $\alpha_0$ is the sum of all concentration parameters. $\psi$ is the digamma function. As a measure for epistemic uncertainty, we take $\frac{N}{\alpha_0}$, where $N$ is the number of classes. 
We evaluate the measures of accuracy and uncertainty of the models on the clean dataset, the dataset with reduced diversity, the dataset with increased sample noise, and the dataset with increased label noise\footnote{For noise generation parameters, please refer to our GitHub page}.  

\subsection{Results}
The results are summarized in Table \ref{tab:uq_methods}. Since uncertainty is quantified differently in RCML than in MCD and DE, their values cannot be directly compared, so the table reports the relative changes in both types of uncertainty with respect to the clean dataset.

As we can observe in the table, in most cases, adding label and sample noises effectively increases the epistemic and aleatoric uncertainties. Interestingly, in most MCD and DE models, the uncertainty decreases when they are trained on data with lower diversity. This may indicate that these approaches fail to recognize data points outside their training distribution, which we will further investigate with the OOD detection task.

We evaluate AUC score for OOD detection based on the epistemic uncertainty. The results are summarized in Table \ref{tab:ood_auc}. We can see that Monte Carlo Dropout and Deep Ensembles fail to provide epistemic uncertainty values suitable for OOD detection in LUMA dataset, with a poor performance of approximately 0.5 AUC value. On the other hand, the RCML achieves an outstanding AUC score of 0.91, indicating that the epistemic uncertainty values quantified with this method can be effectively used for OOD detection.

\begin{figure}
    \centering
    \includegraphics[width=0.8\linewidth]{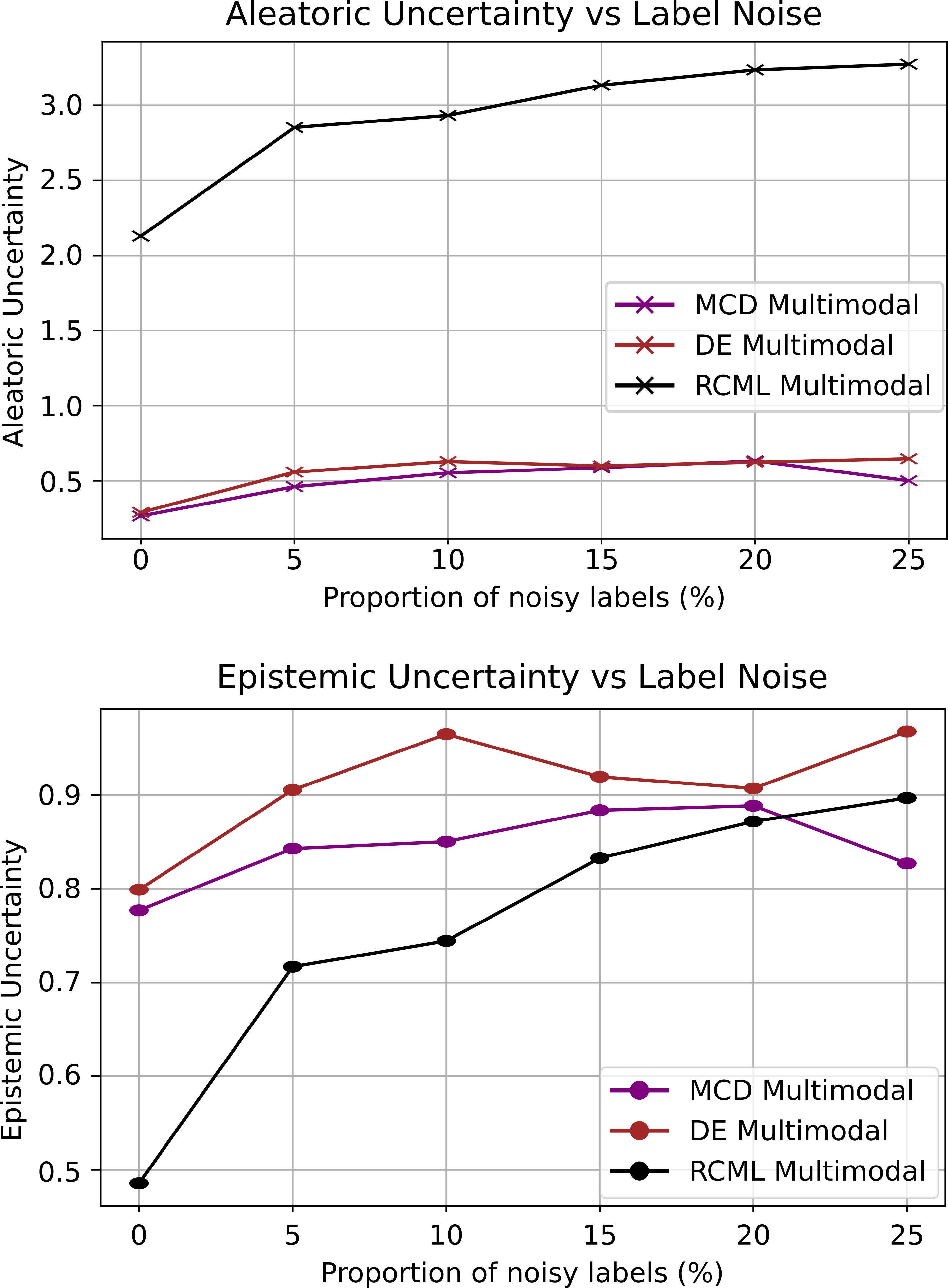}
    \caption{Changes in uncertainty estimations under different proportions of label noise (in percentages). The RCML  approach consistently increases both aleatoric and epistemic uncertainties with increased label noise. In contrast, the MCD  and DE  models sometimes fail to increase the corresponding uncertainty estimations in this experiment.}
    \label{fig:label_noise_vs_unc}
    \Description{Figure 10: Fully described in text and caption.}
\end{figure}

To further evaluate the qualities of the uncertainties of the different models, we estimate the epistemic and aleatoric uncertainties under different amounts of label noise. Ideally, we expect a good uncertainty quantification algorithm to provide higher uncertainty values for more noisy data. As we can see from Figure \ref{fig:label_noise_vs_unc}, only RCML consistently raises the uncertainty estimates under increased label noise, which again shows the higher quality of its uncertainty estimates over the other baselines. 

In Table \ref{tab:accuracy_results}, we present the classification accuracy measures for the clean dataset and variations in accuracy under different types of noise for the three baseline models across each modality. We observe that accuracy always decreases with increasing the label noise, but reducing diversity and increasing sample noise may not always decrease accuracy in the image modality.

In conclusion, the performance of Monte Carlo Dropout and Deep Ensembles indicates a limitation in their suitability for OOD detection in LUMA dataset. This suggests new avenues for further exploratory research to leverage uncertainty estimation for robust detection of out-of-distribution samples. Furthermore, the observed disparities highlight the necessity for a comprehensive benchmarking effort on LUMA dataset, encompassing a broader array of state-of-the-art methodologies.

\section{Conclusion}
\label{sec:conclusion}

In this paper, we propose  \name{}, a multimodal dataset for learning from uncertain and multimodal data. The dataset spans image, audio, and text modalities, and is accompanied by a Python package that allows users to generate customized versions with varying levels and types of noise and uncertainty. To support future research, we also provide a suite of baseline implementations for performance comparison.
The dataset can be easily extended with additional modalities and augmented with more data samples. In the future, we plan to include dependent modalities, enabling more comprehensive studies of uncertainty quantification as well as improved applicability in information retrieval contexts.  The open-source nature of the data compilation pipeline and code for uncertainty and noise generation facilitates the integration of new contributions from the community to promote multimodal uncertainty studies and benchmarking initiatives.

\begin{acks}
    We thank our colleagues and friends for contributing to the audio validation process, and to the HumanSignal team for providing an academic license for Label Studio. 
\end{acks}

\bibliographystyle{ACM-Reference-Format}
\balance
\bibliography{bibliography}





\end{document}